%% file: ms.tex
\documentclass[10pt,twocolumn,letterpaper]{article}

\usepackage[nolist, printonlyused]{acronym}          	
\usepackage{times}
\usepackage{epsfig}
\usepackage{graphicx}
\usepackage{amsmath}
\usepackage{amssymb}
\usepackage{amsthm}
\usepackage{xcolor}
\usepackage[colorinlistoftodos]{todonotes}
\usepackage{ifthen}
\usepackage[utf8]{inputenc}
\usepackage[english]{babel}
\usepackage[backend=biber, style=ieee, maxcitenames=1, mincitenames=1, uniquelist=false]{biblatex}
\usepackage{soul}
\usepackage[shortlabels]{enumitem}
\usepackage{csquotes}
\usepackage{tabularx} 
\usepackage{tikz}
\usetikzlibrary{shadows.blur}

\usepackage[hidelinks,breaklinks=true,bookmarks=false]{hyperref}
\usepackage{cleveref}

\begin{document}

\title{Safety Concerns and Mitigation Approaches Regarding the Use of Deep Learning in Safety-Critical Perception Tasks}

\author{Oliver Willers, Sebastian Sudholt, Shervin Raafatnia, Stephanie Abrecht\\
Robert Bosch GmbH, Chassis Systems Control, Automated Driving\\
74232 Abstatt, Germany
}

\maketitle

\begin{abstract}
    Deep learning methods are widely regarded as indispensable when it comes to designing perception pipelines for autonomous agents such as robots, drones or automated vehicles.
    The main reasons, however, for deep learning not being used for autonomous agents at large scale already are safety concerns.
    Deep learning approaches typically exhibit a black-box behavior which makes it hard for them to be evaluated with respect to safety-critical aspects.
    While there have been some work on safety in deep learning, most papers typically focus on high-level safety concerns.   
    In this work, we seek to dive into the safety concerns of deep learning methods and present a concise enumeration on a deeply technical level.
    Additionally, we present extensive discussions on possible mitigation methods and give an outlook regarding what mitigation methods are still missing in order to facilitate an argumentation for the safety of a deep learning method. 
\end{abstract}

\section{Introduction}
\label{sec:intro}
\input{intro}

\section{Related Work}
\label{sec:relatedwork}
\input{relatedwork}

\section{Definitions}
\label{sec:definitions}
\input{definitions}

\section{Safety Concerns}
\label{sec:safetyconcerns}
\input{safetyconcerns}


\section{Potential Mitigation Approaches}
\label{sec:mitigationapproaches}
\input{mitigationapproaches}

\section{Conclusion}
\label{sec:conclusion}
\input{conclusion}

\section*{Acknowledgments}
Parts of the research leading to the results presented above are funded by the German Federal Ministry for Economic Affairs and Energy within
the project \enquote{Safe AI - Methods and measures for safeguarding AI-based perception functions for automated driving}.
We would like to thank the consortium for the successful cooperation. 
In particular, we would like to thank Peter Schlicht and Christian Hellert for reviewing our work and their thoughtful comments.

\begin{acronym}	
	\acro{AD}{Automated Driving}
	\acro{ADAS}{Advanced Driver Assistance Systems}
	\acro{AI}{Artificial Intelligence}
	\acro{DL}{deep learning}
	\acro{DNN}{Deep Neural Network}
	\acro{E/E}{electrical and/or electronic}	
	\acro{ECE}{expected calibration error}
	\acro{GSN}{Goal Structuring Notation}
	\acro{i.i.d.}{independent and identically distributed}
	\acro{KPI}{key performance indicator}
	\acro{MCE}{maximum calibration error}
	\acro{ML}{machine learning}
	\acro{NN}{Neural Network}
	\acro{CNN}{Convolutional Neural Network}
	\acro{ODD}{Operational Design Domain}
	\acro{ReLU}{Rectified Linear Unit}
	\acro{SOTIF}{Safety of the Intended Functionality}
	\acro{SC}{safety concern}
\end{acronym}

\printbibliography

\end{document}

%% file: intro.tex
During the last years new and exciting applications were enabled by \ac{ML} and especially, by \ac{DL} methods. Their capability of solving problems which cannot be fully specified makes \ac{DL} a key enabler in many applications. 
Therefore, \ac{DL} is also of fundamental importance for the fast growing field of \ac{ADAS} and \ac{AD} as it is not possible to specify an open context in every detail (e.g., the data representation of a pedestrian in all varieties cannot be specified such that it could always be recognized by a rule-based algorithm).

Different from humans, current \ac{DL} algorithms do not learn semantic or causal relationships but simply correlations in data they are presented with.
For example, a \ac{DL} algorithm used for detecting objects in camera images learns correlations between the pixels of the image and object representations, e.g., bounding boxes.
While \ac{DL} algorithms provide state-of-the-art performance, it is more difficult to understand how they arrive at their predictions. 
This poses a problem when releasing systems that incorporate \ac{DL} methods from a safety point of view.
 
While safety related aspects in the automotive area are usually handled through approaches defined in the ISO~26262~\cite{iso_26262}, the usage of \ac{DL} methods introduces a number of additional safety-related aspects not covered in the aforementioned norm.
Most notably, \ac{DL} algorithms may predict incorrect results, e.g., an object detection algorithm may miss to predict an existing object.
These kinds of limitations are not covered in the ISO~26262 but rather in the recently published ISO~PAS~21448 also known as the \ac{SOTIF}~\cite{sotif}.

According to this standard, \ac{SOTIF} is the absence of unreasonable risk due to hazards resulting from functional insufficiencies of the intended functionality.
A prerequisite for achieving \ac{SOTIF} is a proper understanding of the system, its limitations as well as the conditions which may unveil these limitations.
This is a difficult task for systems incorporating \ac{DL} components because the learning process of \ac{DL} algorithms is entirely different from that of a human being.
Humans intuitively analyze systems and their weaknesses on a semantic level, e.g., interpreting a difficult scene as a composition of things like lightning conditions, type and position of objects, behavior of actors, etc.
However, in \ac{DL} the problem space shifts from a semantic level to the level of data representations (e.g., pixel values of an image). 
Thus, \ac{DL}-specific insufficiencies and failure causes are not necessarily intuitive for humans, making it difficult to understand such methods and their limitations. 
Hence, arguing the safety of a system that relies on the correctness of \ac{DL} outputs requires a dedicated safety consideration of such algorithms.

In this paper, we give a concise overview of safety concerns and their underlying problems regarding the use of \ac{DL} algorithms focusing on \acp{DNN}\footnote{Please note that while we focus on \acp{DNN}, a large amount of the safety concerns discussed in this paper may also be valid for other types of \ac{ML}-based methods.}. In particular, we will consider \acp{DNN} used in the perception pipeline of an \ac{ADAS} or \ac{AD} system. Typical use cases for such components are \ac{DNN}-based object detection or semantic segmentation of the input data. The information obtained from these \ac{DNN}-components are then further used in an \ac{ADAS} or \ac{AD} system which may incorporate additional information such as parallel sensing paths or post processing of the \ac{DNN}'s output. The goal of the system is to enable one or multiple functions, e.g., an automated emergency brake or a highway pilot.
Furthermore, we present potential mitigation approaches along with a deep technical discussion.


%% file: relatedwork.tex
The question how one can use \ac{ML} in safety-critical tasks and especially highly automated driving has attracted a considerable amount of research over the last years (e.g., \cite{Kurd2003, Amodei2016, Varshney2016, Burton2017, Salay2018, Gharib2018, Gauerhof2018, Burton2019, Adler2019}). As discussed in the previous section, existing automotive standards such as ISO 26262 do not address the unique characteristics of data-driven approaches used in an open context.

As pointed out in \cite{Gharib2018}, currently, there exists no agreed-upon way to verify and validate \ac{ML} components used in \ac{ADAS} or \ac{AD} systems. In particular, the foundational statistical \ac{ML} principles of empirical risk minimization and average losses are not fully applicable when considering safety, as discussed in \cite{Varshney2016}. However, several works exist which define requirements or safety criteria such a component needs to fulfill.

In \cite{Kurd2003}, the authors derive safety criteria for neural networks from an abstract top-level goal. Thereby, the posed goals and criteria are on a purely functional level outlined in a \ac{GSN}. 
Following this line of work, \textcite{Burton2017} and \textcite{Gauerhof2018} propose a systematic approach using \ac{GSN} in order to argue the safety of \ac{ML}-based components. 
In contrast to \textcite{Kurd2003}, they put the focus already more on the specific issues of \ac{ML} models. In their work, they formulate requirements for an \ac{ML} model derived from the discussed safety concerns. Furthermore, they discuss potential sources of evidences for the constructed assurance case. 
 
In a further work, \textcite{Burton2019} propose an approach for constructing an argumentation for the safety of an \ac{ML} model which they term \emph{performance evidence confidence}. The approach is based on a design-by-contract principle of the safety argumentation which in turn uses safety contracts. These safety contracts provide certain guarantees if a defined set of assumptions hold.

Another work that deals with this topic is given by \textcite{Adler2019}. 
Here, the authors extract areas of activity by a systematic literature search. Based on this, challenges regarding the use of \acp{DNN} in safety critical applications are listed and methods which might help to overcome them are mapped. However, the validity of the list as well as the effectiveness of the mapped methods remains to be shown. 

In this work, we seek to expand the discussion about safety concerns with regard to the usage of \acp{DNN} in safety-critical perception tasks. Furthermore, we concretize these concerns including root causes and discuss potentials as well as limitations of possible mitigation approaches.

%% file: definitions.tex
Before going into the details of safety concerns, we first give definitions for the most crucial terms used in this paper.

A \acf{DNN} is a machine learning model which is made up of layers. The layers may be either connected in a feedforward or recurrent fashion.
Each layer takes some form of data as input, processes it, applies a so-called activation function and then outputs the result.
This output may in turn be used by other layers as input.
The output of the final layer is used as the prediction of the \ac{DNN}.
In most use-cases arising for \acp{DNN} in the context of highly automated driving, the \ac{DNN} is asked to predict a conditional probability $p(Y=y\,|\,X=x)$.
In other words, the \ac{DNN} is tasked to predict the posterior probability for a dependent random variable $Y$ (e.g., class probabilities) based on the independent random variable $X$ (e.g., input images).
For this, one needs to specify the expected type of distribution of $Y$.
This is important as the \ac{DNN} needs to be equipped with a so-called link function which maps to the correct range of $Y$.
If, for example, one wants to perform classification with the \ac{DNN}, $Y$ is typically expected to follow a multinomial distribution.
In this case, the link function of choice is the well-known softmax.
As $X$ and $Y$ are unknown, the typical approach for obtaining a good \ac{DNN} model is to record a dataset $D = \lbrace (x_i, y_i) \rbrace_{i=1}^N$ with realizations of $X$ and $Y$ and perform maximum-likelihood estimation of the parameters with respect to the data.
Here, $x_i$ is a data sample (e.g., camera image) and $y_i$ the corresponding annotation(s) (e.g., bounding boxes of objects to be detected).
In practice, optimization is typically achieved by minimizing the value of the negative log-likelihood function using (stochastic) gradient descent.
The negative log-likelihood function is commonly referred to as loss function in this case.

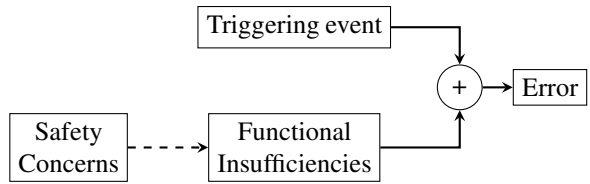
\begin{figure}[t]	
	\centering
	\input{flow_chart_insuff.tex}
	\caption{The relation between safety concerns, functional insufficiencies, triggering event, and error. Concerns potentially lead to insufficiencies inherent in the function. Together with a triggering event, a functional insufficiency provoke an erroneous output of the function.}
	\label{fig:concern_insuff_error}
\end{figure}

According to ISO PAS 21448, \textit{functional insufficiencies} are insufficiencies inherent in the system possibly leading to hazards. Such an insufficiency can appear, e.g., in form of a performance limitation leading to an incomplete or wrong perception of the environment. A functional insufficiency can be unveiled under some conditions. A set of such conditions is referred to as a \textit{triggering event}. In particular, considering a \ac{DNN} module in the perception pipeline of an \ac{ADAS} or \ac{AD} system, such an event can provoke an erroneous output~(see \autoref{fig:concern_insuff_error}) possibly causing a hazardous behavior of the system. 


%% file: flow_chart_insuff.tex
\begin{tikzpicture}

	\node [draw] (a) {Triggering event};
	\node [right of=a,node distance=2.2cm] (b) {};
	\node [draw, circle, below of=b,node distance=0.8cm] (c) {+};
	\node [draw, below of=a,node distance=1.6cm, align=center] (d) {Functional\\Insufficiencies};
	\node [draw, left of=d,node distance=3cm, align=center] (e) {Safety\\Concerns};
	\node [draw, right of=c,node distance=1.2cm] (f) {Error};	
	\draw[thick, dashed, -stealth] (e) -- (d);
	\draw[thick, -stealth] (d) -| (c);
	\draw[thick, -stealth] (a) -| (c);	
	\draw[thick, -stealth] (c) -- (f);
		
\end{tikzpicture}

%% file: safetyconcerns.tex
We define \acp{SC} (or simply concerns) as underlying issues which may negatively affect the safety of a system.
They are either the direct root of a functional insufficiency or describe a black-box-like characteristic of the system which in turn makes it hard to assess safety.
Safety concerns are usually tied to subcomponents of the system.
In particular, there exist specific concerns when deploying a \ac{DL} algorithm in an \ac{ADAS} or \ac{AD} vehicle.
	
The concerns which turn into functional insufficiencies originate from the inherent design of \ac{DL} methods.
In general, a supervised \ac{DL} algorithm tries to extract the joint probability distribution $p(X, Y)$~\cite{Bousquet2004}.
As the distribution is inherently unknown, the only option is to approximate it through a dataset $D$ and extract the characteristics of the distribution from the dataset.
The algorithm produces incorrect results, if its approximation of the underlying distribution $p$ is not good enough at a given data point.

The concerns relating to black-box-characteristics originate from \ac{DL}-specific properties.
\ac{DL} algorithms usually project the input data into high-dimensional spaces which cannot be entirely interpreted by a human anymore.
While it is, for example, well known that classification-based \ac{DL} methods partition their input space into non-convex subspaces, giving semantic meaning to these subspaces is largely impossible.

In the following, we will describe the safety concerns of \ac{DL} algorithms in an \ac{AD} perception pipeline in detail.\\
%
\\
\textbf{Data distribution is not a good approximation of real world \hypertarget{data_dist}{(SC-1)}}
The first overarching concern is that the distribution of the data used in development might not be a good approximation to the one of the real open world which is \emph{a priori} unknown. As mentioned before, the distribution meant here is on the level of data representations, which are high-dimensional and non-intuitive. Therefore, we can only approach them from (or estimate them on) a semantic level by analyzing influencing factors such as daylight, object appearance and weather conditions. This is prone to incompleteness since not all aspects important for the data representation may be covered this way. Besides, the data collection can have other shortcomings which are independent of the level at which it is represented. Examples of such problems are bias (e.g. over- or under-representation of certain factors) or disregarding effects related to different physical deployments (e.g. varying sensor position and angle due to different system variants or manufacturing tolerances). Training and testing a \ac{DNN} with data which do not sufficiently approximate the \ac{ODD} will very likely lead to an insufficient performance or robustness later in the field.\\
\\
\textbf{Distributional shift over time \hypertarget{dist_shift}{(SC-2)}}
A \ac{DNN} is trained and tested at a certain point in time, e.g., during development. 
However, our world is changing continuously. 
This means that even if we would train a \enquote{perfect} algorithm, the probability distribution of the input data will change over time (e.g., new vehicles with a different appearance will be released). 
Since such a change will occur naturally, this concern needs to be addressed by appropriate measures being effective over the product's lifetime.\\
\\
\textbf{Incomprehensible behavior \hypertarget{incomp_behav}{(SC-3)}}
One of the main difficulties in arguing the safety of \acp{DNN} is our inability to explain exactly how they come to a decision. 
In other words, the non-linearity and complexity of \acp{DNN} is a double-edged sword; on the one hand, it enables them to automatically extract features and relate those to outputs via non-linear activation functions, which, in turn, makes them so suitable for solving non-specifiable problems.
On the other hand, those features and their connection to the outputs are rather counterintuitive and incomprehensible for us.
Therefore, unlike in the case of rule-based functions, it is hardly possible to derive a causal relation between the data representation and predictions of the network.
Consequently, identifying weaknesses and failure causes of \acp{DNN} is difficult and sometimes infeasible, impeding the applicability of common safety engineering methods (e.g., fault tree analysis, common cause analysis, etc.).\\
\\
\textbf{Unknown behavior in rare critical situations \hypertarget{rare_crit_sit}{(SC-4)}}
This concern is directly related to the well-known long-tail problem in the context of \ac{AD}. 
The long-tail problem describes the fact that there exists an 
enormous amount of scenarios that have a low occurrence probability.
These scenarios may however be safety-critical.
If one wants to test them, it would require a practically impossible amount of driving hours to capture them by chance.
Regarding this issue, two important aspects need to be mentioned: first, note that according to the statistical learning theory, the performance of an \ac{ML} algorithm evaluated on a test data set can only be generalized if test data, training data and the data which the function is facing later in the open world are \ac{i.i.d.} samples out of the same probability distribution~\cite{Bousquet2004}. Thus, it might be problematic to artificially insert such scenarios in the test data used to estimate the generalization capability of \ac{DNN}'s performance. 
Second, even though one could define a separate dataset in order to test the function with respect to such data, it is hardly possible to identify a rare critical situation from the perspective of a \ac{DNN} \emph{a priori}.
This is due to the fact that \acp{DNN} do not look at semantic content but rather the data itself (see \cref{sec:intro}).
This makes it very difficult to define appropriate test cases in advance in order to deal with this concern.\\
\\
\textbf{Unreliable confidence information \hypertarget{unrel_conf}{(SC-5)}}
In practice, \acp{DNN} will be faced with input data for which they cannot make an accurate prediction. 
This may either stem from an insufficient amount or representativeness of training data or an inherent uncertainty in the data itself (e.g., motion blur). 
Ideally, the \ac{DNN} should reliably indicate if its prediction can be trusted or not. 
This behavior would allow for a number of established safety approaches to be used for a \ac{DNN} component such as giving more weight to parallel information paths, initiating an emergency maneuver or a driver handover.
Most \ac{DNN} algorithms used in practice output some form of posterior probability (e.g., class probabilities in case of a classifier) and one may be tempted to use the value of the highest probability or the information entropy as a measure of confidence. 
This may, however, be highly critical if the probabilities are not well calibrated.
In particular, it has been shown that \acp{DNN} using the standard multinomial cross entropy loss in combination with the softmax as link function tend to be overconfident in their predictions~\cite{Guo2017}.
Even worse, it can be shown that if these \acp{DNN} use \acp{ReLU} as activation functions they can produce arbitrarily high posterior probabilities when dealing with data far away from the training data~\cite{Hein2019}.
While confidence information may not benefit the solution of the problem itself, it serves as a vital enabler of a safety argument in a respective safety case.
For example, well calibrated confidence information may be used in a multi-sensor system to fuse concurrent predictions from different sensors.\\
\\
\textbf{Brittleness of \acp{DNN} \hypertarget{brittle_nn}{(SC-6)}}
As shown by many works, the brittleness of \acp{DNN} is a major safety concern. 
This includes the robustness against common perturbations such as noise or certain weather conditions, e.g.~\cite{Hendrycks2019}, and translations/rotations, e.g.~\cite{Alcorn2018}, as well as targeted perturbations known as adversarial examples, e.g.~\cite{Szegedy2014, Goodfellow2015}. 
Note that regarding adversarial examples, the so-called adversarial patches are of special interest in the context of \ac{ADAS} and \ac{AD} (e.g., ~\cite{Eykholt2018, NirMorgulis2019, Lee2019}).
This is due to the fact that a would-be attacker can simply change the operation environment of a vehicle instead of having to hack into the vehicle itself.
Physical adversarial patch-based attacks do thus scale considerably better than those based on overlaying the raw sensor data recorded in a vehicle with noise.\\
\\
\textbf{Inadequate separation of test and training data \hypertarget{data_sep}{(SC-7)}} 
Another concern is that test data might be inadequately separated from training data. For training and testing \acp{DNN}, the data is usually divided into training, validation and test datasets. In order not to overestimate the \ac{DNN}'s performance, the test dataset needs to be (sufficiently) uncorrelated to the other ones. However, in practice, highly correlated data is usually acquired because, e.g., data is recorded in sequences (i.e. consecutive frames are rather similar) or data is recorded at the same locations several times.
Another aspect is that developers tend to optimize on test datasets during training because they strive for the maximum performance which is measured on this test data. Therefore, a training process is continued until performance goals of a network are met on the test dataset.
Good and labeled data is expensive and thus, rare in practice, but using a test dataset several times means also an optimization with respect to the test data leading to an overestimation of a \ac{DNN}'s performance.\\
%
\\
\textbf{Dependence on labeling quality \hypertarget{label_q}{(SC-8)}}
In the case of supervised learning, labeled datasets are required for training and testing a \ac{DNN}. Notice that the labeling, which is typically done manually, and its quality directly affect the resulting function and therefore, the obtained test results as shown, e.g., in \cite{Haase-Schutz2019}.
In particular, if the label quality is not sufficient, the results obtained during testing may be misleading. As a result, the function could have an insufficient performance later in the field.
Hence, the labeling quality needs to be ensured in order to argue the safety of such a learning function.\\
\\
\textbf{Insufficient consideration of safety in metrics \hypertarget{insuff_metric}{(SC-9)}}
Using state-of-the-art metrics such as mean average precision and false positive/negative rate, only the \textit{average} performance of \acp{DNN} is evaluated. 
Additionally, when assessing the performance of a \ac{DNN}, typically all elements of a test dataset influence the performance metric.
There may, however, be elements which the \ac{DNN} predicted incorrectly but would not impact the system itself.
For example, consider the case of a \ac{DNN} used for pedestrian detection which serves the function of an automated emergency brake.
If the car is driving at 30\,kph and fails to detect a pedestrian at 500\,m distance, this will in all likelihood not have an impact on the safety of the system.
However, in common metrics, such a person will be counted in the same way as a person standing directly in front of the car.
This will inevitably lead to giving the \ac{DNN} a worse safety rating than is actually the case.\\



%% file: mitigationapproaches.tex
Releasing an \ac{ADAS} or \ac{AD} system requires a comprehensive argumentation to show that all concerns related to the system's safety are identified, understood and mitigated. 
After having discussed the safety concerns regarding the use of \acp{DNN} within such systems in \cref{sec:safetyconcerns}, we present several promising mitigation approaches (MAs) which could be used in order to provide supporting arguments and evidences for a safety case.\\
\\
\textbf{Well-justified data acquisition strategy (\hypertarget{m_data_strat}{MA-1})} The basis for testing \ac{ML} functions is an appropriate dataset reflecting the context in which the function is supposed to work. 
In particular, one needs to argue that the dataset used is a suitable representation of the data which the \ac{DNN} will be faced within the \ac{ODD}. 
As pointed out before, the distribution which is relevant here is on the level of the data representations (e.g., pixel-level distribution).
Finding suitable random samples from this distribution is - in most cases - highly non-trivial, mainly due to the dimensionality of the data.
Thus, we propose to follow a two-step approach here.
%
The first step is to specify the data content, as well as the data acquisition and selection process, in a structured and thorough manner. 
For this, essential \ac{ODD} factors such as weather conditions, road types, occurring objects as well as their variations in the \ac{ODD} need to be determined, see e.g., {\cite{Koopman2019}}.
Additional factors such as tolerances in the mounting positions of the sensors and predictable changes over the product’s lifetime (e.g. sensor aging) should be considered as well. 
Finally, the existence of specified variations and their frequencies in the acquired data need to be verified.

The aforementioned analysis happens on a semantic level and may not fully cover the specifics of the data at hand (e.g., certain biases in the pixel distribution of an image).
Thus, the second step is to analyze the raw data and find suitable datapoints which are missing from the first step.
This can, for example, be achieved by finding a latent representation of the data using a variational autoencoder and sampling the latent space in a suitable manner.\\
\\
\textbf{Enabling the output of reliable confidence information (\hypertarget{m_conf_output}{MA-2})}
As explained before, the posterior probability predicted by a \ac{DNN} tends to be overconfident even for inputs close to the training data \cite{Guo2017} and may be arbitrarily high when moving away from the training data \cite{Hein2019}.
In order to be able to output reliable confidence information, a number of approaches have been proposed.
In \cite{Guo2017} a number of heuristic approaches are evaluated, which either make use of the logit or posterior probability outputs in order to calibrate the output probabilities and in turn allow them to be used as a reliable measure of confidence.

Besides heuristics, other approaches have made use of Bayesian methods in order to extract uncertainties.
In \cite{Gal2016} the authors use dropout during inference which turns their neural network into a Bayesian model with the weights being represented by Bernoulli distributions.
They show that when dropout is used at inference time, one approximately marginalizes over the weights of the neural network using Monte Carlo integration.
This approach is hence termed \emph{Monte Carlo Dropout}.
Another Bayesian approach is presented by \textcite{Blundell2015}.
Here, the authors model the weights of the neural network using Gaussian distributions and minimize the ELBO loss.
They achieve this using also Monte Carlo integration to approximately marginalize over the weights of the neural network.

Besides the actual method itself, it is still an open question how one can determine if a measure of confidence is reliable or not in the context of \ac{AD}.
In \cite{Naeini2015} the use of \ac{ECE} and \ac{MCE} is proposed.
Both metrics operate on the probabilities predicted from the neural network.
First, the maximum posterior probability is quantized into a desired number of bins for a test dataset.
Then, the accuracy is computed for each. Generally, the outputs are well-calibrated if the accuracy of each bin is equal to the average probability in this bin.
The difference in these two values is called calibration error.
While for \ac{ECE} the calibration error is averaged over all bins, \ac{MCE} simply returns the largest calibration error.
However, a main drawback of both \ac{ECE} and \ac{MCE} is that both metrics depend on a parameter, namely the number of bins.
This parameter heavily influences the obtained result.\\
\\
\textbf{Using gray-box methods (\hypertarget{m_gray_box}{MA-3})}.
A major impediment to the safety argumentation of \acp{DNN} is their black-box character~\hyperlink{incomp_behav}{SC-3}. 
Even though turning the black-box to a white-box will be scarcely possible in the foreseeable future, several methods were introduced recently to gain understanding of the root causes for \ac{DNN}'s predictions by visualizing decisive parts of the input (e.g., gradient-weighted class activation mapping~\cite{Selvaraju2017}) or by forcing the \ac{DNN} to provide more interpretable outputs (e.g., object attributes~\cite{Lampert2014}). 
While these methods cannot enable an analytical safety evaluation, they still can contribute to a safety case, e.g., by making the analysis of a test result more meaningful or by supporting the extraction of uncertainties for \ac{DNN}'s prediction (e.g., by analyzing the distribution of decisive parts of an image with respect to certain object classes). Note that the trustworthiness of such methods needs to be shown which is indeed a non-trivial task as well.\\
\\
\textbf{Specification of adversarial threat models and incorporation of defense methods (\hypertarget{m_adv_def}{MA-4})}
Before being able to defend against adversarial examples, one must first determine a threat model, which in essence represents an assumption on what a possible attacker is capable to perform as an attack.
Most of the current work in adversarial examples focuses on data-level threat models, meaning that an attacker is allowed to change the values of a given datum.
For example, in computer vision-based problems this is typically achieved by changing the values of pixels in an image at arbitrary locations.
This kind of threat model typically involves some form of budget that may not be exceeded, e.g., the difference in pixel values between an original image and an adversarial example may not exceed a certain amount, oftentimes measured in either $l_2$ or $l_\infty$ norm (e.g., \cite{Madry2018,Goodfellow2015,Metzen2017}).
Other data-level threat models include adversarial patches~\cite{Brown2017} or affine transformation-based attacks~\cite{Engstrom2019}.
Of course, allowing data-level changes may oftentimes be an unrealistic or highly improbable threat model.
For example, in the case of autonomous vehicles, an attacker would need access to the pixel buffer in order to alter pixel values.
This form of attack does not scale well and is thus probably a neglectable threat model.
However, there exist a number of techniques, which are known as physical adversarial examples, that do scale well.
Here, the environment in which a datum is recorded is altered instead of the datum itself.
Common techniques for this include sticker-based attacks which can either be applied to objects in the environment (e.g.,~\cite{Eykholt2018,Eykholt2018a,Lee2019}) or be used to partially occlude the sensor which is used for recording the data (e.g.,~\cite{Li2019}).

There exist many other threat models which have not been listed so far\footnote{For a concise overview of common threat models see, e.g., \cite{Yuan2019}.}.
In general, there exists no model which may be assumed by default.
In the future, there might be standards and norms which define an appropriate model for a given domain (e.g., physical based attacks for \ac{AD}).
However, in the meantime the choice of threat model must be made on a per-case basis and argued accordingly.

Having chosen and argued for a specific threat model, one has to deploy defense mechanisms which protect against falling victim to adversarial examples.
The main problem with most known defense mechanisms is that they may have given good results initially but were quickly exposed after having been published.
This has been the fate of distillation-based defenses~\cite{Papernot2016} (exposed in~\cite{Carlini2017}), defenses based on transforming the input such as JPEG compression~\cite{Guo2018} (exposed in~\cite{Athalye2018}) and gradient-obfuscation methods~\cite{Buckman2018} (exposed in~\cite{Athalye2018}).
As of writing this paper, there only exist two approaches for defending against adversarial examples, which are effective to at least a certain degree and are somewhat accepted in the \ac{ML} community\footnote{Defending against adversarial examples is currently a heavily researched topic and there may exist other effective methods.}.
First, there is an empirical approach known as adversarial training with PGD adversaries~\cite{Madry2018}.
This method tries to optimize a \ac{DNN} to predict the correct class for a given sample's strongest adversarial example.
While this approach is not able to guarantee that it actually finds the strongest adversary under a given threat model, it is very flexible with respect to the model actually used.
For example, the commonly used bounded pixel-level threat model can be easily replaced by other models such as rotation- or sticker-based attacks.
The second approach uses a convex outer approximation of reachable activations of the \ac{ReLU} units of a neural network to defend against adversarial examples~\cite{Wong2018a}.
This method can give guaranteed lower bounds on the loss values of adversarial examples.
A drawback of this analytic method is that the training procedure takes considerably longer than standard SGD training\footnote{There have been improvements in the training time of this method in order to scale to larger datasets, see \cite{Wong2018}.}.

Beside making the network itself more robust, other approaches aim at detecting adversarial attacks, e.g., by using a trained subnetwork~\cite{Metzen2017}. Even though the \ac{DNN} would still be fooled by the attack in this case, the information that the \ac{DNN}'s prediction is not trustworthy at this moment could trigger an appropriate system reaction preventing harm. However, such a detector-network could be attacked as well which means that its robustness needs to be argumented too.\\
\\
\textbf{Testing (\hypertarget{m_test}{MA-5})}
Naturally, a key component of a safety argumentation is testing usually including verification and validation activities. While verification rather addresses issues which are already known or foreseeable (e.g., lack of robustness against certain perturbations), validation  focuses on identifying unknown issues.
In the following, we will refer to mitigation approaches that address these issues as MA-5a and MA-5b respectively.

\textbf{MA-5a:} Known or predictable critical cases can be assessed via targeted testing. This approach supports mitigating \hyperlink{data_dist}{SC-1}, \hyperlink{rare_crit_sit}{SC-4} and \hyperlink{brittle_nn}{SC-6}. The selection of test data is key for a thorough analysis of \acp{DNN}. One of the methods for identifying targeted test cases is HAZOP (Hazard\&Operability, \cite{Kletz1986}). It is a standard safety procedure used to systematically identify malfunctions and risks of a complex system. 
In \cite{Zendel2015}, the authors adapt HAZOP to computer vision systems and provide a catalog containing an extensive set of known critical situations for computer vision tasks as a basis for assessing the quality and thoroughness of test data.
Of special interest is the stability of \ac{DL} algorithms with respect to certain effects in the input space (e.g. blur, windscreen smudges or exposure related effects).
As highlighted by \textcite{Zendel2018}, the evaluation of robustness requires a targeted addition of difficult samples into a test dataset. A benchmark for robustness against known corruptions and perturbations is introduced in \cite{Hendrycks2019}.
Another approach for effectively testing \ac{DNN} algorithms is search-based-testing~\cite{Zhang2019}. This technique aims at exploring the input space in a targeted manner enabling, e.g., a sensitivity analysis with respect to certain \ac{ODD} factors or different combinations of them.
Note that while some of the approaches mentioned can make use of real data (recorded on public roads or test tracks) others require artificially generated data. 
Thus, it is important to mention that for obtaining reliable test results on synthetic data, the validity of this data with respect to real data has to be shown. This is, in turn, a highly non-trivial problem.\footnote{Even though synthetic data may look \enquote{realistic} to a human, the data-level distribution may be significantly different leading to non-meaningful test results.}.

\textbf{MA-5b:} The unknown and unpredictable problems associated with deploying \acp{DNN} in a safety-critical open-world context can only be identified \textit{by chance}. For this purpose, field test data need to be collected randomly in accordance to the guidelines mentioned in \hyperlink{m_data_strat}{MA-1}.  

Such a testing mainly addresses \hyperlink{rare_crit_sit}{SC-4}, but also supports the mitigation of \hyperlink{brittle_nn}{SC-6}, by providing a means for finding previously unknown safety-critical situations
\footnote{It is important to note that for reasons described in \hyperlink{data_sep}{SC-7}, the test set used for the ultimate performance evaluation needs to remain unseen until final testing.}. 
Note that the open-context nature of the operational domain, renders the coverage of the entire problem space via brute-force approaches practically infeasible. Instead, one needs to combine field testing with other methods, as pointed out in this paper, to enable the release of such systems.\\
\begin{table*}[!h]
	\renewcommand{\arraystretch}{1.3}
	\caption{Overview of safety concerns and associated mitigation approaches.}
	\label{tab:concerns_mitigation_mapping}
	\centering
	\begin{tabularx}{\textwidth}[ht!]{ >{\hsize=0.73\hsize}X | >{\hsize=1.27\hsize}X}	
		\hline	
		\textbf{Safety concern} & \textbf{Mitigation approaches} \\
		\hline\hline
		Data distribution is not a good approximation of real world \hyperlink{data_dist}{(SC-1)} & Well-justified data acquisition strategy \hyperlink{m_data_strat}{(MA-1)}, enabling the output of reliable confidence information \hyperlink{m_conf_output}{(MA-2)}, testing \hyperlink{m_test}{(MA-5)}, deep analysis of test results obtained in an iterative development process \hyperlink{m_test_anal}{(MA-6)}, labeling guidelines \hyperlink{m_label_guid}{(MA-8)}\\
		\hline	
		Distributional shift over time \hyperlink{dist_shift}{(SC-2)} & Enabling the output of reliable confidence information \hyperlink{m_conf_output}{(MA-2)}, continuous learning and updating \hyperlink{m_cont_learn}{(MA-10)}\\
		\hline
		Incomprehensible behavior \hyperlink{incomp_behav}{(SC-3)} & Using gray-box methods \hyperlink{m_gray_box}{(MA-3)}\\
		\hline
		Unknown behavior in rare critical situations \hyperlink{rare_crit_sit}{(SC-4)} & Well-justified data acquisition strategy \hyperlink{m_data_strat}{(MA-1)}, enabling the output of reliable confidence information \hyperlink{m_conf_output}{(MA-2)}, testing \hyperlink{m_test}{(MA-5)}, deep analysis of test results obtained in an iterative development process \hyperlink{m_test_anal}{(MA-6)}, continuous learning and updating \hyperlink{m_cont_learn}{(MA-10)}\\
		\hline
		Unreliable confidence information \hyperlink{unrel_conf}{(SC-5)} & Enabling the output of reliable confidence information \hyperlink{m_conf_output}{(MA-2)}, using gray-box methods \hyperlink{m_gray_box}{(MA-3)}, testing \hyperlink{m_test}{(MA-5)}\\
		\hline
		Brittleness of \acp{DNN} \hyperlink{brittle_nn}{(SC-6)} & Enabling the output of reliable confidence information \hyperlink{m_conf_output}{(MA-2)}, specification of adversarial threat models and incorporation of defense methods \hyperlink{m_adv_def}{(MA-4)}, testing \hyperlink{m_test}{(MA-5)}, deep analysis of test results obtained in an iterative development process \hyperlink{m_test_anal}{(MA-6)}, continuous learning and updating \hyperlink{m_cont_learn}{(MA-10)}\\
		\hline
		Inadequate separation of test and training data \hyperlink{data_sep}{(SC-7)} & Data partitioning guidelines \hyperlink{m_data_part}{(MA-7)} \\
		\hline
		Dependence on labeling quality \hyperlink{label_q}{(SC-8)} & Labeling guidelines \hyperlink{m_label_guid}{(MA-8)}\\
		\hline			
		Insufficient consideration of safety in metrics \hyperlink{insuff_metric}{(SC-9)} & Evaluating performance with respect to safety \hyperlink{m_perf_safe}{(MA-9)}\\
		\hline		
	\end{tabularx}
\end{table*}
\\
\textbf{Deep analysis of test results obtained in an iterative development process (\hypertarget{m_test_anal}{MA-6})}.
As is known, \ac{DL} is a data-driven approach and its development should be pursued in an iterative way. Discovered weaknesses of the \ac{DL} component are continuously mitigated by optimizing architectures and hyperparameters or by adding new data that covers previously missing aspects. Hence, a fundamental part of this process is analyzing the intermediate results, ideally leading to a continuous improvement. In order to extract as much information as possible from these results, the analysis should be performed in a structured, careful, and if possible, automated manner (e.g., by extracting systematic weaknesses from comprehensive metadata by which the data should be enriched beforehand). In addition to cases where the \ac{DL} component makes wrong predictions, cases associated with high uncertainty should be considered. This is important because even though the function might have been correct \enquote{at random}, it could lead to wrong predictions and therefore, cannot be ignored. This approach can contribute to the mitigation of \hyperlink{rare_crit_sit}{SC-4} and \hyperlink{brittle_nn}{SC-6}.\\
\\
\textbf{Data partitioning guidelines (\hypertarget{m_data_part}{MA-7})} In order to address \hyperlink{data_sep}{SC-7} and estimate a \ac{DNN}'s performance correctly, guidelines regarding partitioning the data into training, validation and test datasets are necessary. 
In particular, test data must not be correlated with training data since otherwise the generalization capability of the \ac{ML} algorithm will be overestimated. 
This means that, e.g., consecutive frames of a video sequence may not be assigned to different partitions. 
Further measures could be that test data needs to be acquired at different days and locations as training data. 
Such guidelines need to be well-justified and the partitioning needs to be subsequently reviewed with regard to the guidelines.\\
\\
\textbf{Labeling guidelines (\hypertarget{m_label_guid}{MA-8})}
The dependence of supervised learning methods on well-labeled data (see \hyperlink{label_q}{SC-8} in \cref{sec:safetyconcerns}) requires strict labeling guidelines and checks. 
The guidelines should be defined with respect to the specific task (e.g. semantic segmentation or object detection) and should ideally contain additional application-specific annotations in order to enable an automated evaluation, e.g., of the relative frequencies of \ac{ODD} factors such as weather conditions, object-specific metadata, etc. Guidelines compilation has to be justified and the adherence to them needs to be reviewed. Appropriately performed, this mitigates \hyperlink{label_q}{SC-8} and supports the argumentation with respect to \hyperlink{data_dist}{SC-1}.\\
\\
\textbf{Evaluating performance with respect to safety (\hypertarget{m_perf_safe}{MA-9})}.
As pointed out above, current state-of-the-art performance metrics in machine learning are calculating average values not considering safety with respect to a certain function (e.g., automated emergency brake)~\hyperlink{insuff_metric}{SC-9}. 
Realizing that it will not be possible to reach 100\% performance, it is obvious that a safety argumentation is hardly possible based on these metrics. However, considering an object detection component in the perception of an \ac{AD} vehicle, it is actually not necessary to assure that all objects are detected but all the objects which are \textit{relevant} with respect to system safety. Additionally, one could further refine that all \textit{relevant} objects need to be detected \textit{or} a low confidence value needs to indicate that the \ac{DNN} might be wrong such that the system can manage the situation safely (e.g., by relying more on other information paths). 
Another important aspect is the analysis of errors over time. If one considers, for example, an object detection network, missing an object in one single frame might not be problematic at all because this can be compensated, e.g., by state-of-the-art object tracking methods or by plausibility checks (e.g., a pedestrian will probably not disappear within a few milliseconds). But if an object is not detected in several consecutive frames, the severity of the error is much higher.
Therefore, tailored evaluation metrics are necessary in order to meaningfully assess \acp{DNN} from a safety perspective.\\
\\
\textbf{Continuous learning and updating (\hypertarget{m_cont_learn}{MA-10})}
In order to maintain the safety of a \ac{DNN}-based component, the open context and distributional shift over time problems (issued in \hyperlink{rare_crit_sit}{SC-4} and in \hyperlink{dist_shift}{SC-2} respectively) need to be addressed in the product's life cycle. In particular, the \ac{DNN} could face novel inputs in which the parameter distribution (e.g. pixel values in an image) differ from that of the data seen during development. This can occur either because the difference oversteps the generalization abilities of the network (long-tailed open context) or the input includes something completely new (e.g. a new type of vehicle) which has not existed before (temporal distributional shift) possibly leading to hazards. Therefore, it may be necessary to continually develop the algorithm further and updating it.
Note that continuous learning does not necessarily mean online learning of the \ac{DNN} already applied in the vehicle.
While this approach is generally possible, it comes with its own specific problems, namely continuous validation of the newly learned model in the car with only minimal computation power as well as weak to no supervision.
Continuous learning as proposed here includes an offline development step. New and useful data is recognized by a \ac{DNN} or some other mechanism and send back to an offline data center where a new version of the \ac{DNN} is trained and validated.
Finally, the old \ac{DNN} in the \ac{ADAS} or \ac{AD} vehicle is replaced with the new one, either through software-over-the-air solutions or in a workshop.
This process ensures the in-use \ac{DNN} to be up-to-date while still having the ability to make use of large scale computation power for validation.

%
%
%
%
%
%

%% file: conclusion.tex

In this work, we have presented a concise list of safety concerns regarding deep learning methods used in perception pipelines of autonomous agents, especially highly automated vehicles.
We also presented an extensive discussion on possible mitigation approaches addressing those safety concerns (the mapping is presented in \cref{tab:concerns_mitigation_mapping}).
It is important to note that the discussed approaches have very different maturity and complexity. Furthermore, while all of the approaches can definitely contribute to a safety case, 
for the time being it remains an open question when a specific safety concern is sufficiently mitigated.
In particular, many of the mitigation methods involve parameters for which there does not exist a single \textit{correct} value.
For example, some methods supply a \ac{KPI} telling the user how well the deep learning algorithm under test performed with respect to this \ac{KPI}.
However, the threshold for this \ac{KPI} used to determine whether the deep learning algorithm is safe cannot be obtained analytically in many cases.
Thus, it is essential to collect knowledge and consolidate this in standardization activities in order to define suitable processes, practices and thresholds.

%% file: ms.bib
@InProceedings{Burton2017,
  author    = {Burton, Simon and Gauerhof, Lydia and Heinzemann, Christian},
  title     = {Making the {Case} for {Safety} of {Machine} {Learning} in {Highly} {Automated} {Driving}},
  booktitle = {Computer {Safety}, {Reliability}, and {Security}},
  year      = {2017},
  pages     = {5--16},
}

@Article{Amodei2016,
  author  = {Amodei, Dario and Olah, Chris and Steinhardt, Jacob and Christiano, Paul F. and Schulman, John and Mané, Dan},
  title   = {Concrete {Problems} in {AI} {Safety}},
  journal = {arXiv},
  year    = {2016},
}

@Article{Guo2017,
  author   = {Guo, C. and Pleiss, G. and Sun, Y. and Weinberger, K. Q.},
  title    = {On {Calibration} of {Modern} {Neural} {Networks}},
  journal  = {arXiv},
  year     = {2017},
}

@InProceedings{Blundell2015,
  author    = {Blundell, Charles and Cornebise, Julien and Kavukcuoglu, Koray and Wierstra, Daan},
  title     = {Weight {Uncertainty} in {Neural} {Networks}},
  booktitle = {International Conference on Machine Learning},
  year      = {2015},
  pages     = {1613--1622},
}

@InProceedings{Wong2018,
  author    = {Wong, E. and Schmidt, F. and Hendrik Metzen, J. and Zico Kolter, J.},
  title     = {Scaling Provable Adversarial Defenses},
  booktitle = {{arXiv}},
  year      = {2018},
}

@InCollection{Bousquet2004,
  author    = {Bousquet, Olivier and Boucheron, Stéphane and Lugosi, Gábor},
  title     = {Introduction to {Statistical} {Learning} {Theory}},
  booktitle = {Advanced {Lectures} on {Machine} {Learning}: {ML} {Summer} {Schools} 2003},
  year      = {2004},
  pages     = {169--207},
}

@InProceedings{Alcorn2018,
  author    = {Alcorn, Michael A. and Li, Qi and Gong, Zhitao and Wang, Chengfei and Mai, Long and Ku, Wei-Shinn and Nguyen, Anh},
  title     = {Strike (with) a {Pose}: {Neural} {Networks} {Are} {Easily} {Fooled} by {Strange} {Poses} of {Familiar} {Objects}},
  booktitle = {{arXiv}},
  year      = {2018},
}

@inproceedings{Hendrycks2019,
	title={{Benchmarking Neural Network Robustness to Common arXivuptions and Perturbations}},
	author={Dan Hendrycks and Thomas Dietterich},
	booktitle={International Conference on Learning Representations},
	year={2019},
}

@InProceedings{Eykholt2018a,
  author    = {Eykholt, Kevin and Evtimov, Ivan and Fernandes, Earlence and Li, Bo and Rahmati, Amir and Xiao, Chaowei and Prakash, Atul and Kohno, Tadayoshi and Song, Dawn},
  title     = {Robust {Physical}-{World} {Attacks} on {Deep} {Learning} {Models}},
  booktitle = {{Computer} {Vision} and {Pattern} {Recognition}},
  year      = {2018},
}

@InProceedings{Eykholt2018,
  author    = {Eykholt, Kevin and Evtimov, Ivan and Fernandes, Earlence and Li, Bo and Rahmati, Amir and Tramèr, Florian and Prakash, Atul and Kohno, Tadayoshi and Song, Dawn},
  title     = {Physical {Adversarial} {Examples} for {Object} {Detectors}},
  booktitle = {{arXiv}},
  year      = {2018},
}

@InProceedings{Zendel2015,
	author={O. {Zendel} and M. {Murschitz} and M. {Humenberger} and W. {Herzner}},
	booktitle={IEEE International Conference on Computer Vision},
	title={{CV-HAZOP: Introducing Test Data Validation for Computer Vision}},
	year={2015},
	volume={},
	number={},
	pages={2066-2074},
}

@article{NirMorgulis2019,
  author = {{Nir Morgulis} and {Alexander Kreines} and {Shachar Mendelowitz} and {Yuval Weisglass}},
  title  = {Fooling a {Real} {Car} with {Adversarial} {Traffic} {Signs}},
  year   = {2019},
  journal = {arXiv},
}

@InProceedings{Haase-Schutz2019,
	author={C. {Haase-Schütz} and H. {Hertlein} and W. {Wiesbeck}},
	booktitle={Intelligent Vehicles Symposium},
	title={{Estimating Labeling Quality with Deep Object Detectors}},	
	year={2019},
	volume={},
	number={},
	pages={33-38},
}

@inproceedings{Madry2018,
	title={Towards Deep Learning Models Resistant to Adversarial Attacks},
	author={Aleksander Madry and Aleksandar Makelov and Ludwig Schmidt and Dimitris Tsipras and Adrian Vladu},
	booktitle={International Conference on Learning Representations},
	year={2018},
}

@InProceedings{Burton2019,
  author    = {Burton, Simon and Gauerhof, Lydia and Sethy, Bibhuti Bhusan and Habli, Ibrahim and Hawkins, Richard},
  title     = {Confidence {Arguments} for {Evidence} of {Performance} in {Machine} {Learning} for {Highly} {Automated} {Driving} {Functions}},
  booktitle = {Computer {Safety}, {Reliability}, and {Security}},
  year      = {2019},
  pages     = {365--377},
}

@InProceedings{Gharib2018,
  author={M. {Gharib} and P. {Lollini} and M. {Botta} and E. {Amparore} and S. {Donatelli} and A. {Bondavalli}},
  booktitle={International Conference on Dependable Systems and Networks Workshops},
  title={{On the Safety of Automotive Systems Incorporating Machine Learning Based Components: A Position Paper}},
  year={2018},
  volume={},
  number={},
  pages={271-274},
}

@article{Varshney2016,
	title={{Engineering Safety in Machine Learning}},
	journal={Information Theory and Applications Workshop},
	publisher={IEEE},
	author={Varshney, Kush R.},
	year={2016},
}

@Article{Zhang2019,
  author  = {Zhang, Jie M. and Harman, Mark and Ma, Lei and Liu, Yang},
  title   = {Machine {Learning} {Testing}: {Survey}, {Landscapes} and {Horizons}},
  journal = {arXiv},
  year    = {2019},
}

@InProceedings{Szegedy2014,
  author    = {Szegedy, Christian and Zaremba, Wojciech and Sutskever, Ilya and Bruna, Joan and Erhan, Dumitru and Goodfellow, Ian and Fergus, Rob},
  title     = {Intriguing properties of neural networks},
  booktitle = {International {Conference} on {Learning} {Representations}},
  year      = {2014},
}

@InProceedings{Goodfellow2015,
  author    = {Goodfellow, Ian and Shlens, Jonathon and Szegedy, Christian},
  title     = {Explaining and {Harnessing} {Adversarial} {Examples}},
  booktitle = {International {Conference} on {Learning} {Representations}},
  year      = {2015},
}

@Article{Lee2019,
  author  = {Lee, Mark and Kolter, J. Zico},
  title   = {On {Physical} {Adversarial} {Patches} for {Object} {Detection}},
  journal = {arXiv},
  year    = {2019},
}

@InProceedings{Hein2019,
	author    = {Matthias Hein and Maksym Andriushchenko and Julian Bitterwolf},
	title     = {{Why ReLU networks yield high-confidence predictions far away from the training data and how to mitigate the problem}},
	booktitle = {Computer Vision and Pattern Recognition},
	year      = {2019},
	pages 	  = {41--50},
}

@inproceedings{Koopman2019,
	title = {How many operational design domains, objects, and events?},
	booktitle = {Workshop on {AI} {Safety}},
	author = {{Koopman, P.} and {Fratrik, F.}},
	year = {2019}
}

@inproceedings{Naeini2015,
author = {Pakdaman Naeini, Mahdi and Cooper, Gregory and Hauskrecht, Milos},
year = {2015},
pages = {2901-2907},
title = {{Obtaining Well Calibrated Probabilities Using Bayesian Binning}},
booktitle = {Conference on Artificial Intelligence}
}

@article{Adler2019,
	title = {Hardening of {Artificial} {Neural} {Networks} for {Use} in {Safety}-{Critical} {Applications} - {A} {Mapping} {Study}},
	journal = {arXiv},
	author = {Adler, Rasmus and Akram, Mohammed Naveed and Bauer, Pascal and Feth, Patrik and Gerber, Pascal and Jedlitschka, Andreas and Jöckel, Lisa and Kläs, Michael and Schneider, Daniel},
	year = {2019}
}

@incollection{Kurd2003,
	address = {Berlin, Heidelberg},
	title = {Establishing {Safety} {Criteria} for {Artificial} {Neural} {Networks}},
	volume = {2773},
	booktitle = {Knowledge-{Based} {Intelligent} {Information} and {Engineering} {Systems}},
	publisher = {Springer Berlin Heidelberg},
	author = {Kurd, Zeshan and Kelly, Tim},
	editor = {Goos, Gerhard and Hartmanis, Juris and van Leeuwen, Jan and Palade, Vasile and Howlett, Robert J. and Jain, Lakhmi},
	year = {2003},
	pages = {163--169}
}

@InProceedings{Gal2016,
  title = 	 {{Dropout as a Bayesian Approximation: Representing Model Uncertainty in Deep Learning}},
  author = 	 {Yarin Gal and Zoubin Ghahramani},
  booktitle = 	 {International Conference on Machine Learning},
  pages = 	 {1050--1059},
  year = 	 {2016},
  editor = 	 {Maria Florina Balcan and Kilian Q. Weinberger},
  volume = 	 {48},
}

@inproceedings{Salay2018,
	author={Salay, Rick and Queiroz, Rodrigo and Czarnecki, Krzysztof},
	title={An Analysis of ISO 26262: Machine Learning and Safety in Automotive Software},
	booktitle={WCX World Congress Experience},
	publisher={SAE International},
	year={2018},
}

@article{Brown2017,
	title={{Adversarial Patch}},
	author={Tom B. Brown and Dandelion Man\'{e} and Aurko Roy and Mart\'{i}n Abadi and Justin Gilmer},
	year={2017},
	journal = {arXiv},
}

@article{Yuan2019,
	title={{Adversarial Examples: Attacks and Defenses for Deep Learning}},
	volume={30},
	number={9},
	journal={Transactions on Neural Networks and Learning Systems},
	publisher={Institute of Electrical and Electronics Engineers (IEEE)},
	author={Yuan, Xiaoyong and He, Pan and Zhu, Qile and Li, Xiaolin},
	year={2019},
	pages={2805–2824}
}

@InProceedings{Engstrom2019,
	title = 	 {{Exploring the Landscape of Spatial Robustness}},
	author = 	 {Engstrom, Logan and Tran, Brandon and Tsipras, Dimitris and Schmidt, Ludwig and Madry, Aleksander},
	booktitle = 	 {International Conference on Machine Learning},
	pages = 	 {1802--1811},
	year = 	 {2019},
}

@article{Li2019,
	title={Adversarial camera stickers: A physical camera-based attack on deep learning systems},
	author={Juncheng Li and Frank R. Schmidt and J. Zico Kolter},
	year={2019},
	journal = {arXiv},
}

@inproceedings{Papernot2016,
	title={{Distillation as a Defense to Adversarial Perturbations Against Deep Neural Networks}},
	booktitle={Symposium on Security and Privacy},
	author={Papernot, Nicolas and McDaniel, Patrick and Wu, Xi and Jha, Somesh and Swami, Ananthram},
	year={2016},
}

@article{Carlini2017,
  title={Towards Evaluating the Robustness of Neural Networks},
  author={Nicholas Carlini and David A. Wagner},
  journal={IEEE Symposium on Security and Privacy},
  year={2017}  
}

@inproceedings{Guo2018,
	title={{Countering Adversarial Images using Input Transformations}},
	author={Chuan Guo and Mayank Rana and Moustapha Cisse and Laurens van der Maaten},
	booktitle={International Conference on Learning Representations},
	year={2018},
}

@inproceedings{Athalye2018,
	author={Anish Athalye and Nicholas Carlini and David A. Wagner},
	title={{Obfuscated Gradients Give a False Sense of Security: Circumventing Defenses to Adversarial Examples}},
	year={2018},
	pages={274-283},
	booktitle={International Conference on Machine Learning},
}

@inproceedings{Buckman2018,
	title={{Thermometer Encoding: One Hot Way To Resist Adversarial Examples}},
	author={Jacob Buckman and Aurko Roy and Colin Raffel and Ian Goodfellow},
	booktitle={International Conference on Learning Representations},
	year={2018},
}

@inproceedings{Wong2018a,
	author    = {Eric Wong and J. Zico Kolter},
	title     = {{Provable Defenses against Adversarial Examples via the Convex Outer
	Adversarial Polytope}},
	booktitle = {International Conference on Machine Learning},
	pages     = {5283--5292},
	year      = {2018},
}

@INPROCEEDINGS{Selvaraju2017,
	author={R. R. {Selvaraju} and M. {Cogswell} and A. {Das} and R. {Vedantam} and D. {Parikh} and D. {Batra}},
	booktitle={International Conference on Computer Vision},
	title={Grad-CAM: Visual Explanations from Deep Networks via Gradient-Based Localization},
	year={2017},
	pages={618-626},	
}

@ARTICLE{Lampert2014,
	author={C. H. {Lampert} and H. {Nickisch} and S. {Harmeling}},
	journal={Transactions on Pattern Analysis and Machine Intelligence},
	title={{Attribute-Based Classification for Zero-Shot Visual Object Categorization}},
	year={2014},
	volume={36},
	number={3},
	pages={453-465},	
}

@book{Kletz1986,
	title={{HAZOP \& HAZAN: Notes on the Identification and Assessment of Hazards}},
	author={Kletz, Trevor A.},
	series={Hazard Workshop Modules},
	year={1986},
	publisher={Institution of Chemical Engineers}
}

@InProceedings{Zendel2018,
author = {Zendel, Oliver and Honauer, Katrin and Murschitz, Markus and Steininger, Daniel and Fernandez Dominguez, Gustavo},
title = {WildDash - Creating Hazard-Aware Benchmarks},
booktitle = {European Conference on Computer Vision},
year = {2018}
}

@inproceedings{Metzen2017,
  title={On Detecting Adversarial Perturbations},
  author={Metzen, Jan Hendrik and Genewein, Tim and Fischer, Volker and Bischoff, Bastian},
  booktitle = {Proceedings of 5th International Conference on Learning Representations},
  year={2017}  
}

@inproceedings{Gauerhof2018,	
	title = {Structuring {Validation} {Targets} of a {Machine} {Learning} {Function} {Applied} to {Automated} {Driving}},
	booktitle = {Computer {Safety}, {Reliability}, and {Security}},
	publisher = {Springer International Publishing},
	author = {Gauerhof, Lydia and Munk, Peter and Burton, Simon},
	editor = {Gallina, Barbara and Skavhaug, Amund and Bitsch, Friedemann},
	year = {2018},
	pages = {45--58}
}

@misc{iso_26262,
	title = {Road vehicles - functional safety ({ISO} 26262)},
	publisher = {International Standards Organisation (ISO)},
	author = {{International Standards Organisation (ISO)}},
	year = {2018}
}

@misc{sotif,
	title = {Road vehicles — Safety of the intended functionality ({ISO/PAS} 21448)},
	publisher = {International Standards Organisation (ISO)},
	author = {{International Standards Organisation (ISO)}},
	year = {2019}
}
